\documentclass{article}

\usepackage{amsmath}
\usepackage{PRIMEarxiv}

\usepackage[utf8]{inputenc} 
\usepackage[T1]{fontenc}    
\usepackage{hyperref}       
\usepackage{url}            
\usepackage{booktabs}       
\usepackage{amsfonts}       
\usepackage{nicefrac}       
\usepackage{microtype}      
\usepackage{graphicx}
\graphicspath{{media/}}     
\title{GLIP: Electromagnetic Field Exposure Map Completion by Deep Generative Networks
\thanks{\textit{{
This work has been accepted to the IEEE PIMRC 2024 for publication. Copyright may be transferred without notice, after which this version may no longer be accessible.}}: 
}}

\author{
{Mohammed Mallik\textsuperscript{2}, Davy P. Gaillot\textsuperscript{1}, 
and Laurent Clavier\textsuperscript{1,2}} \\\\
\small
\textsuperscript{1}Univ. Lille, CNRS, UMR 8520 - IEMN, F-59000, Lille, France (e-mail: firstname.name@univ-lille.fr)\\
\textsuperscript{2}IMT Nord Europe, France (e-mail: firstname.name@imt-nord-europe.fr)\\
}
\begin{document}
\maketitle

\begin{abstract}
In Spectrum cartography (SC), the generation of exposure maps for radio frequency electromagnetic fields (RF-EMF) spans dimensions of frequency, space, and time, which relies on a sparse collection of sensor data, posing a challenging ill-posed inverse problem. Cartography methods based on models integrate designed priors, such as sparsity and low-rank structures, to refine the solution of this inverse problem. In our previous work, EMF exposure map reconstruction was achieved by Generative Adversarial Networks (GANs) where physical laws or structural constraints were employed as a prior, but they require a large amount of labeled data or simulated full maps for training to produce efficient results. In this paper, we present a method to reconstruct EMF exposure maps using only the generator network in GANs which does not require explicit training, thus overcoming the limitations of GANs,
such as using reference full exposure maps. This approach uses a prior from sensor data as Local Image Prior (LIP) captured by deep convolutional generative networks independent of learning the network parameters from images in an urban environment. Experimental results show that, even when only sparse sensor data are available, our method
can produce accurate estimates. 
\end{abstract}


\section{Introduction}
In our daily lives, wireless communication systems have seamlessly integrated themselves, becoming an indispensable element. Therefore, the monitoring of phenomena associated with wireless systems, such as radio-frequency electromagnetic field (RF-EMF) exposure, has great importance. Within an urban environment, many electromagnetic field (EMF) sources, encompassing WiFi, 2G, 3G, 4G, and the latest 5G mobile technologies, contribute to the electromagnetic medium.

While 5G holds the promise of substantial improvements over its predecessors\cite{7338410, RePEc, gajvsek2015electromagnetic}, apprehensions about its implementation have been growing. 
Regulatory organizations, including the International Commission on Non-Ionizing Radiation Protection (ICNIRP), the Institute of Electrical and Electronics Engineers (IEEE), and the World Health Organization (WHO) have undertaken extensive research on establishing human exposure standards for EMFs. This is imperative as mobile devices and base stations emitting EMFs for radio communication must adhere to regulatory human exposure levels \cite{international2020guidelines, bailey2019synopsis}. Therefore, the effects
of exposure to wireless systems need to be monitored.
Despite advances in this field, the wide variety of
parameters, e.g., materials, sensor arrangement, or environment topology, make monitoring of EMF exposure relatively complex and often require a large
number of measurements, which is both time-consuming and expensive. 
While standard kriging interpolation methods \cite{matheron1963principles} are frequently employed, neural network based techniques often exhibit outstanding performance for RF map estimation. \cite{teganya2020data,wang2020sensor,wang2022prediction}.

In \cite{s20010311}, the authors employed a GAN \cite{goodfellow2014generative} to estimate the power spectrum map in urban cognitive radio networks. In their study, 
the generation of power spectrum maps is accomplished through a deep learning regression task, and these maps are subsequently translated into a color scale.  The authors exclusively relied on the inverse polynomial law model to generate the reference full maps for training to estimate power spectrum maps (PSM). Zhuo et al. \cite{8794603} employed a self-supervised approach, leveraging GANs, to synthesize a comprehensive RF map of a designated region from a sparsely sampled RF dataset. Integral to their methodology, they utilized the K-nearest neighbor algorithm to generate the reference maps for training. In \cite{mallik2022towards,mallik2022eme, mallik2023eme}, authors inferred EMF exposure by deep learning models in indoor and urban environments, specifically UNet \cite{ronneberger2021u} based convolutional neural network (CNN) and conditional GAN (cGAN) \cite{mirza2014conditional}, respectively. In all these works \cite{mallik2022towards,mallik2022eme, mallik2023eme}, to train the deep learning models, reference full maps were generated from ray-tracing simulators \cite{egea2019vehicular,amiot2013pylayers}. This training phase raises a major issue: simulated maps are not perfect and require time and expertise to be generated. Training also requires a long time and computational resources (GPU, memory, runtime), which makes a deployment in new areas challenging. Matrix completion, natural image priors, and the ability to infer complex information based on sparse representations of the input data have been demonstrated by autoencoders \cite{shi2014light}, and they appear perfectly suited to address our problem \cite{daribo2010depth,shi2014light,long2019low,szczypkowski20243d,layer2015pet, buades2005review}.

In this work, we propose a method using only the Generator Network (noted as Generative Local Image Prior - GLIP) based on encoder-decoder architecture which does not require a large training set or reference full maps to reconstruct EMF exposure. This method aims to meet the important requirements for spatial interpolation, namely, the absence of assumptions about the nature of the data distribution, the very limited number of sensor measurements as observed data, and the computational complexity (memory, GPU, runtime). In statistical learning as proposed in \cite{buades2005review,szczypkowski20243d}, the encoder-decoder model can reconstruct images from the remaining pixels without a conventional learning process. Our proposed model is designed to reconstruct EMF exposure accurately when using sparse sensor measurements as feature, named as called Local Image Prior (LIP). The model yields better results compared to random features in GAN networks \cite{goodfellow2014generative}. The paper is organized as follows. In Section \ref{s1}, we describe the methodology that includes problem formulation, dataset, proposed method, and model details. Experimental results are presented in Section \ref{s3}. The conclusion is given in Section \ref{s4}.

\section{Methodology} 
\label{s1}
\subsection{Inverse problem formulation}
In this work the aim is to reconstruct the RF-EMF exposure map within a 1 $km^2$ rectangular region, discretized into an $M \times N$ grid situated in Lille, France. The reconstruction relies solely on sparsely distributed fixed sensors randomly positioned within the area. Each sensor, located at coordinates $(m,n)$ within the grid, where $m\in\{1,\cdots,M\}$ and $n\in\{1,\cdots,N\}$, is capable of locally measuring the exposure $e_{(m,n)} \in \mathbb{R}^K$, where $K$ is the number of sensors. The task of estimating the exposure $e \in \mathbb{R}^{M\times N}$ at each grid point can be viewed as an inverse problem. The objective of EMF exposure reconstruction is to ascertain a function $f_\theta:\mathbb{R}^K\rightarrow	\mathbb{R}^{M \times N}$ capable of predicting exposure values in locations where sensors are not available. Here, $\theta$ denotes the parameters of the function $f$. Various modeling approaches, such as kriging, or statistical learning, can be employed to derive this function.

\subsection{Exposure Reconstruction by Deep Learning}

\begin{figure}[t]
\centering
\includegraphics[scale=0.3]
{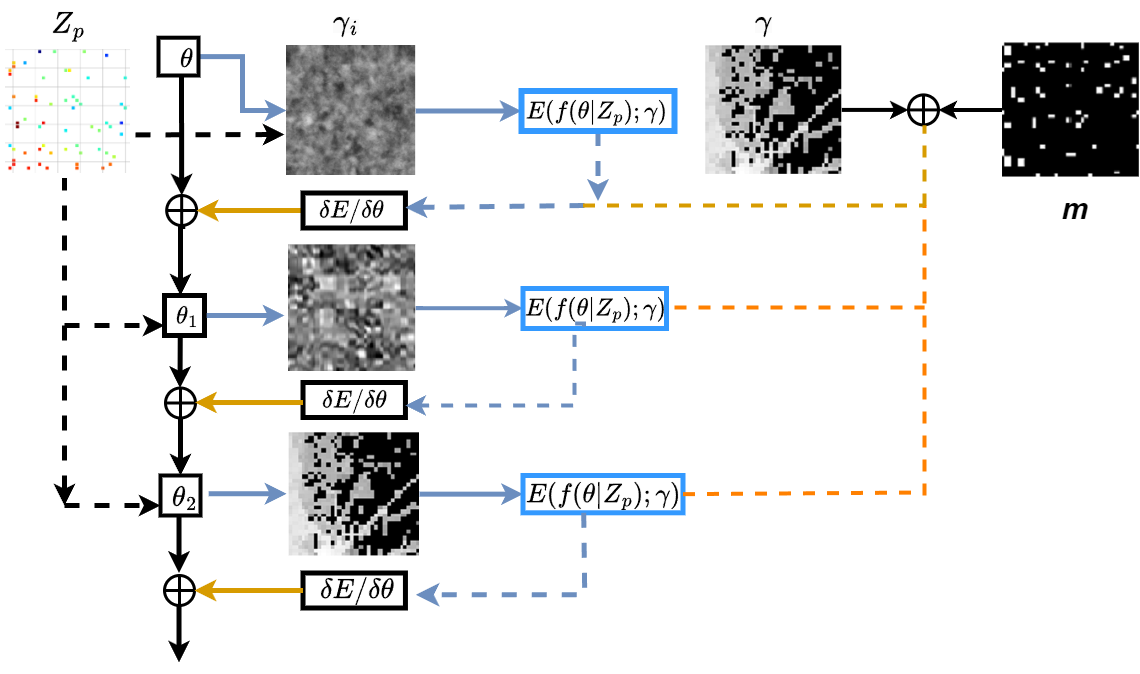}
    \caption{The exposure reconstruction GLIP process. $Z_p$ is the input of the network. $\theta$ is the set of parameters and $\gamma_0$ the initial output of the network. $\gamma$ and $m$ are the reconstructed and mask images. $E$ is the loss function corresponding to each features. $f$ is the network function.}
    \label{fig:fig7}
\end{figure}

In the proposed method, the exposure map image $\gamma$ which needs to be reconstructed is defined by Eq. \eqref{eq:1} as described in \cite{teganya2020data}:
\begin{equation}\label{eq:1}
    \gamma = f(\theta|Z_p),
\end{equation}
where $f$ denotes the deep generative network function, $\theta$ is the set of parameters of the neural network and $Z_p$ is the input of the network and image prior. In this work, the selected prior is the sparse exposure image $\gamma_0$ as the local image prior, LIP. The objective of the model is to minimize the difference between the reconstructed exposure image $\gamma$ and the sparse exposure image $\gamma_0$ to predict the missing values. This makes it an optimization problem as given in Eq.\eqref{eq:2}:
\begin{equation}\label{eq:2}
    \underset{\theta}{\mathop{\arg \min}} \; ||\gamma - \gamma_0||.
\end{equation}
Substituting Eq. \eqref{eq:1} in Eq. \eqref{eq:2}, we use ADAM optimizer to solve the problem presented in Eq. \eqref{eq:3}:
\begin{equation}\label{eq:3}
    \theta_{opt} = \mathop{\arg \min}_\theta E(f(\theta|Z_p); \gamma),
\end{equation}
In the context of exposure reconstruction, the objective function $E(f(\theta|Z_p); \gamma)$ is only calculated from the observed values in $\gamma$. Hence, a binary mask $m \in \{0,1\}^{M\times N}$ is used to take into account only the points observed in the sparse exposure image $\gamma$. $\gamma_i$ denotes the reconstructed map at iteration $i$ during the training. hence, the objective function becomes: 
\begin{equation}\label{eq:4}
    E(\gamma,\gamma_i) = ||(\gamma-\gamma_i) \odot m||^2.
\end{equation}
The norm 
in \eqref{eq:4} will be the squared error between the observed and predicted values. 
The $\odot$ represents the hardmard's product or element-by-element multiplication. A visual representation of our method is depicted in Fig. \ref{fig:fig7}. We underline that only the measured values in $\gamma$ are needed to realize the training of the generative network.

\subsection{Generator Architecture}

\begin{figure}[h]
\centering
\includegraphics[scale=0.6]{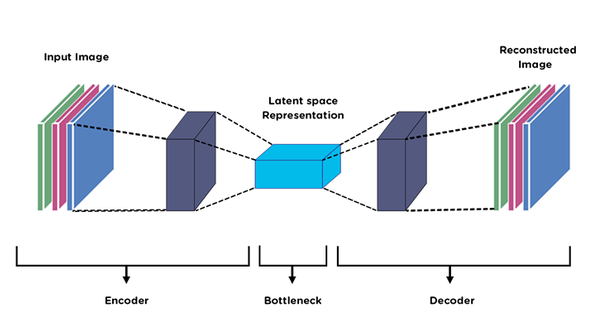}
\caption{Proposed GLIP Generator Network}
\label{fig:cae}
\end{figure}

Fig. \ref{fig:cae} illustrates the CNN architecture used in this paper,
which was adapted from \cite{ronneberger2021u}. 
The popular U-net architecture \cite{ronneberger2021u} is modified such that the skip connections contain a convolutional layer. The decoder uses a down-sampling and upsampling-based scaling-expanding structure, which makes the effective receptive field of the network increase as the input goes deeper into the network. The convolution layers, each followed by a batch normalization function and Leaky Rectified Linear unit (LeakyReLU) function, act as a filter with a kernel that moves over the entire field. Besides, the skip connection enables the later layers to reconstruct the feature maps with both local details and global texture. Here, the input $Z_p$ can be initialized with a fixed tensor of the same spatial size as $\gamma$ filled with random noise. The proposed framework can deal with grayscale images or any other coding of colored images. 
In the dimension reduction or encoder part, we use a CNN. The transfer of
information via the kernel between each convolution layer is well suited to spatial pattern learning. The encoder is constructed with 6 layers to recover complex spatial relationships between observed points.
The second part, called the decoder, aims to increase the dimension until $\gamma_i$ is obtained. The decoder consists of a series of up-convolutional layers or transposed convolutions. These layers upscale the feature maps to match the size of the corresponding feature maps from the encoder. Specifically, in the decoder with skip connections, the input from the encoder's bottleneck layer is first up-sampled to match the dimensions of the original input. Then, this up-sampled feature map is concatenated with the feature maps from earlier layers of the encoder, effectively reintroducing high-level semantic information lost during dimensionality reduction. In our model, we have used 6 skip connections.
The addition of skip connections is very detrimental to the efficiency of the approach. We have used nearest neighbor up-sampling in our method. This can reproduce each pixel according to its nearest neighbor without modification.

Indeed, unlike the experiments carried out in \cite{buades2005review} on 80\% observed data, we limit ourselves to 1\%.  The proposed model's details are given in Table \ref{tab:table_mse}:
\begin{table}[!htp]
\caption{CNN parameters.}
\centering
\begin{tabular}{|l|c|}
\hline
\multicolumn{1}{|c|}{Parameters} & \multicolumn{1}{c|}{Value} \\ \hline
Optimizer                              &  ADAM - Adaptive Moment \\\hline

Kernel Size                                &  2, 3, 4     \\\hline
Non-Linearity                           & LeakyReLU \\ \hline
Upsampling                               &  Nearest Neighbour   \\\hline
Downsampling                               &  Strided convolution2D   \\\hline
Loss function & Mean Squared Error\\\hline
Epochs & 150 \\\hline
\end{tabular}
\label{tab:table3}
\end{table}

\section{Results}\label{s3}

To evaluate our method, different scenarios are considered. The performance is investigated with different numbers of measurement sensors: 20, 40, 60 and 100 spread in the 1 km$^2$ area. We tested the neural model for two cases, random input (generated from a uniform distribution) - GRIP (Generative Random Input Prior) and GLIP. 

\subsection{Evaluation protocol}

\subsubsection{Evaluation metrics}
To evaluate the performance of our system, two summarized metrics are used. First, mean square error (MSE) is given by:
\begin{equation}
 \mathrm {MSE} ={\frac {1}{n}}\sum _{i=1}^{n}\left(Y_{i}-{\hat {Y_{i}}}\right)^{2},
\end{equation}
where $(Y_i -\hat{Y_i})$ denotes the error between the reference value $Y_i$ and the predicted value $\hat{Y_i}$ and $n$ is the number of points in the image.

We also use the mean absolute error (MAE), given by:
\begin{equation}
    \mathrm {MAE} =\frac{1}{n}\sum _{i=1}^{n}\left|Y_{i}-\hat{Y}_{i}\right|,
\end{equation}
which limits the impact of large errors on the resulting error metric.


\subsection{Implementation Details and Dataset}

\begin{figure}[htbp]
\centering
\includegraphics[scale=0.3]{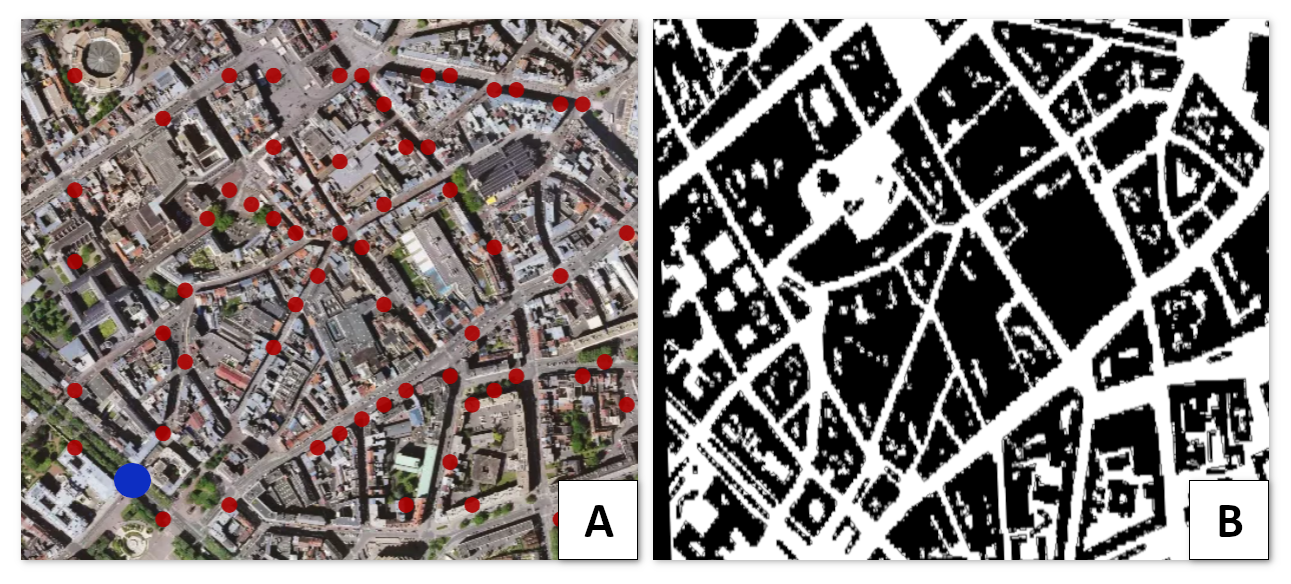}
\caption{Visualization of the region of interest. \textbf{a)} showcases the experimental 1$km^2$ area, where red and blue represent sensors and transmitter respectively. \textbf{b)} reveals the city topology extracted as a raster from OpenStreetMap.}
\label{fig:fig2}
\end{figure}

Our region of interest (RoI) is a 1 $km^2$ area in Lille city center. 
We establish a grid of sensor locations at a height of 1.5 meters, visually depicted in Fig. \ref{fig:fig2}.

\begin{figure}[h]
\centering
\includegraphics[width=\columnwidth]{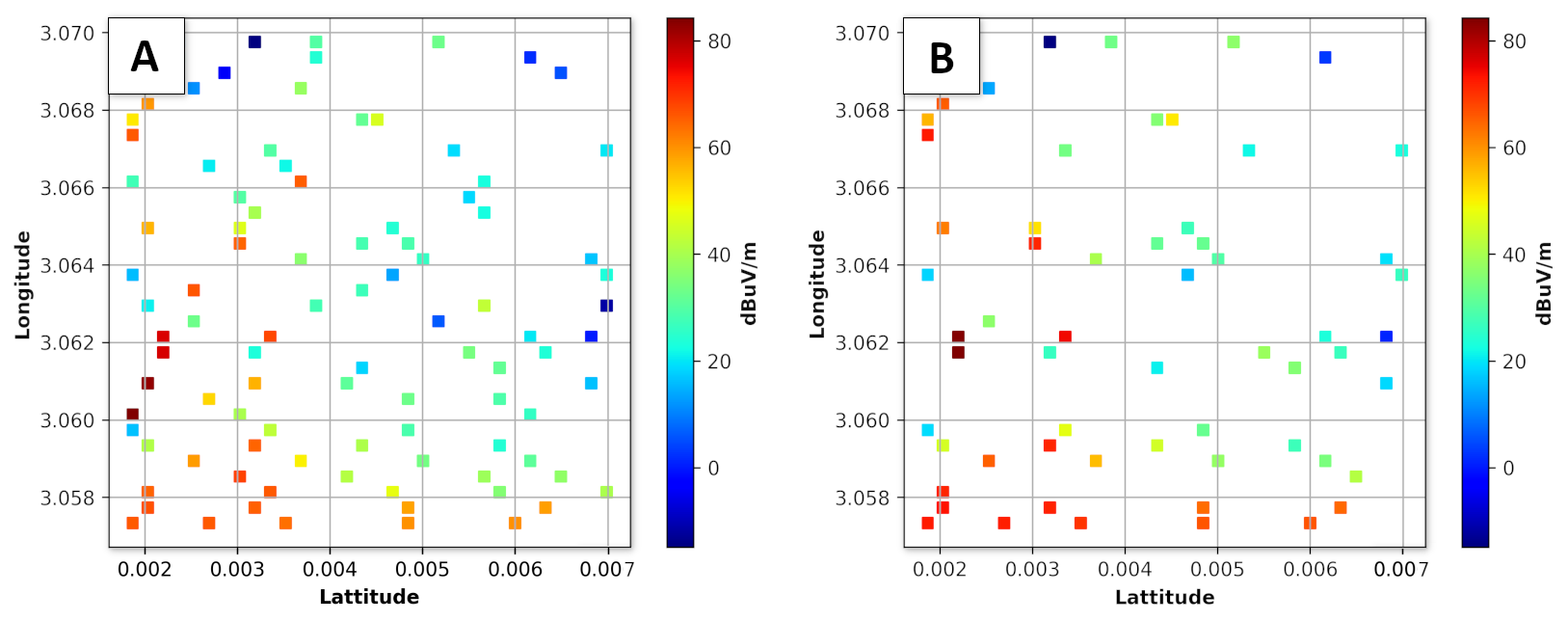}
\caption{Visualization of the LIP \textbf{a)} LIP - 100 sensors , \textbf{b)} LIP - 60 sensors  }
\label{fig:fig3}
\end{figure}
Using Veneris \cite{egea2019vehicular} a 3D ray tracing simulator, one or two transmitters are positioned within this RoI to provide exposure data for the sensors with sparsity increments of 20, 40, 60, and 100 sensors. The transmit power is 120 $W$ and the transmission frequency is 5.89 $GHz$. Moreover, the environment (roads, buildings, etc.) has been taken from OpenStreetMap to incorporate the city topological effect in the proposed method by supressing the building locations in the predicted map $\gamma$. The dataset consists of matrices, where $M$ and $N$ denote the number of rows and columns of the map grid, and each picture is $\gamma \in \mathbb{R}^{M \times N}$. Figure \ref{fig:fig3} provides an example of a sparse sensor map $\gamma$ and LIP.

\begin{figure*}[htbp]
    \centering
\includegraphics[width=\textwidth]{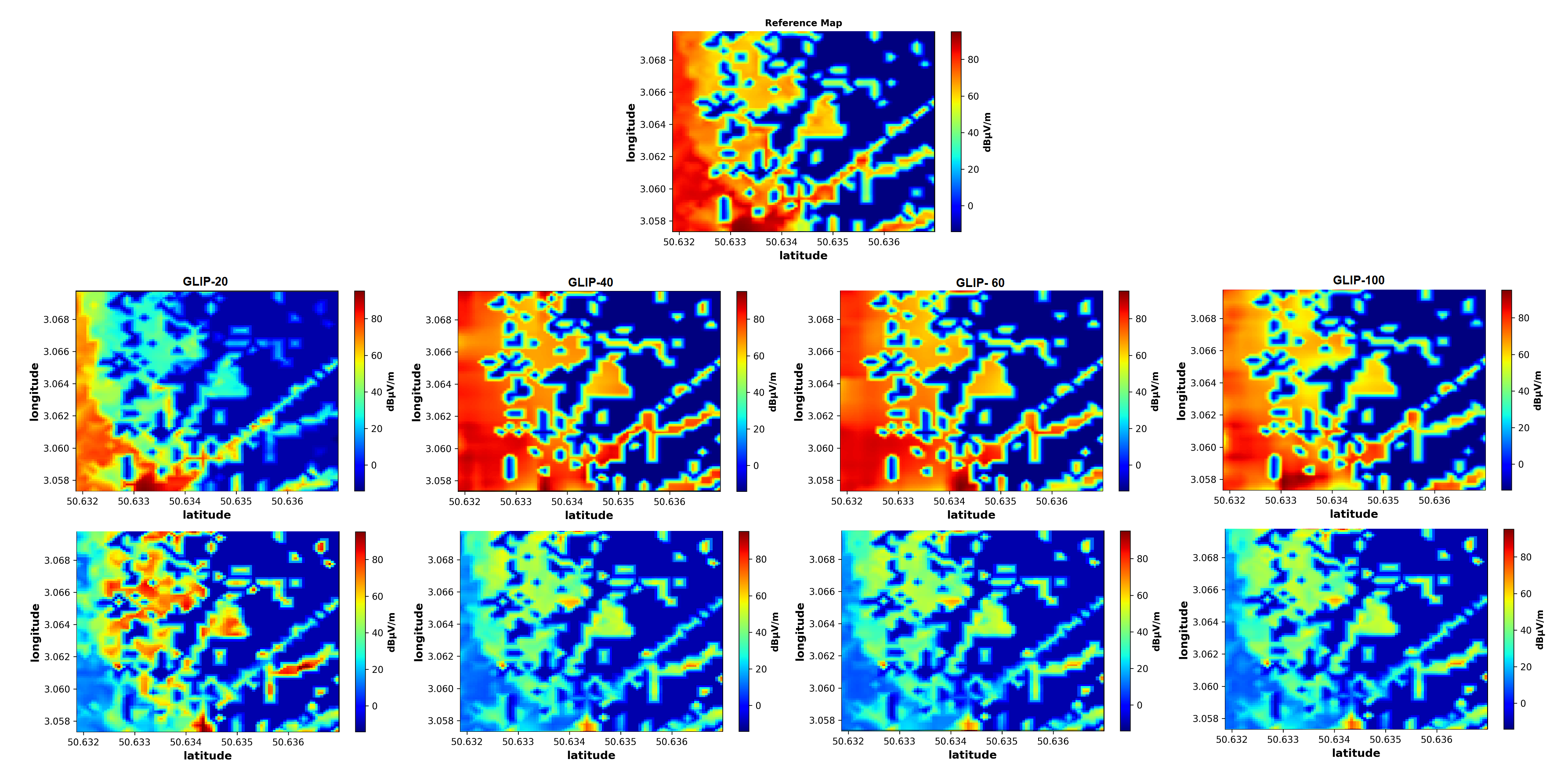}
\caption{Experimental results for GLIP: 1st row: Reference map from Veneris, 2nd row (Left to right): reconstructed maps for 20, 40, 60 and 100 sensors, 3rd row (Left to right): corresponding error maps.}
    \label{fig:res}
\end{figure*}

\subsection{Visual Analysis}
The GLIP method's reconstructed maps are illustrated
in Figure \ref{fig:res}. 
With only 20 measurement points, the reconstruction falls short and fails to represent the reference map. The failure of the method to accurately reconstruct exposure from 20 sensors is evident in the error map, where a significant concentration of errors is observed at the center of the grid. This occurrence can be attributed to the absence of available data points or sensors in that particular area, resulting in inaccuracies in the reconstruction process. While with an increased number of measurement points from 40 to 100, our model performs better reconstruction quality and a low MSE of 2,7$.10^{-5}$ when 100 sensors are used for reconstruction as seen in 
Fig. \ref{fig:res} and the low errors in the error map.

\subsection{Quantitative Analysis and Impact of Sensor Density}
We present in Table \ref{tab:table_mse} and \ref{tab:table_mae} the MSE and MAE of the predicted and actual values of exposure. 

\begin{table}[htbp]  
\centering
\caption{MSE of the estimated exposure values}\label{tab:table_mae}
\label{tab:table_mae}
\begin{tabular}{cccc} 
\toprule
Number of sensors & GLIP & GRIP \\ 
\midrule
20 / $km^2$  & 3,789.$10^{-5}$ & 1,058.$10^{-2}$ \\ 
40 / $km^2$  & 3,041.$10^{-5}$ & 2,427.$10^{-3}$\\
60 / $km^2$  &2,845.$10^{-5}$ & 
6,120.$10^{-4}$ \\
100 / $km^2$ & 2,679.$10^{-5}$ & 
1,736.$10^{-4}$ \\ \bottomrule
\end{tabular}
\label{tab:table_mse}
\end{table}

\begin{table}[htbp]
\caption{MAE of the estimated exposure values}\label{tab:table_mae}
\centering
\begin{tabular}{cccc} 
\toprule
Number of sensors & GLIP & GRIP \\ 
\midrule
20 / $km^2$  & 3,754.$10^{-3}$ & 
5,761.$10^{-2}$\\ 
40 / $km^2$  & 3,271.$10^{-3}$ & 2,145.$10^{-2}$\\
60 / $km^2$  & 2,957.$10^{-3}$ & 9,570.$10^{-2}$\\
100 / $km^2$ & 2,946.$10^{-3}$ & 6,301.$10^{-3}$\\ 
\bottomrule
\end{tabular}
\label{tab:table_mae}
\end{table}

As expected, as shown in Table \ref{tab:table_mse}, increasing the number of sensors reduces the MSE, leading to better reconstruction performance for both GLIP and GRIP. 
However, when using GRIP, the MSE is significantly higher than GLIP (for 100 sensors MSE - GRIP: 1,74.$10^{-4}$ and GLIP: 2,68.$10^{-5}$) proving that using a LIP as random input leads to more accurate results. It is to be noted that the maximum exposure was  0.101 $V/m$ in the dataset. The main conclusion is that the density of sensors and how they are distributed along with using an appropriate prior play a significant role in the reconstruction process. 

In Table \ref{tab:table_mae} below, the MAE of predicted and actual exposure gives the same tendency as the MSE. By increasing the number of sensors from 20 to 100, the reconstruction MAE falls as low as 2,95.$10^{-3}$ for 100 sensors leading to an accurate estimation of exposure. 

Additional investigation is imperative to fine-tune sensor placement. The choice of the number of sensors to be deployed depends on the error that can be accepted and the cost of the deployment. Another constraint that needs to be studied in more detail (but is use-case dependent) is that the location of the sensors is limited to specific places, such as lampposts, and they must not be accessible to people.  


\section{Conclusion}\label{s4}
The study introduces GLIP, a method leveraging generative modeling for estimating RF-EMF exposure maps in an urban area. Unlike 
previously proposed methods \cite{mallik2022eme,mallik2022towards}, GLIP employs a 
generator architecture which enables exposure maps to be accurately predicted without the need for prior learning and a large training dataset of complete reference maps.
Remarkably, the method demonstrates its effectiveness in accurately predicting maps with a minimum of input data, encompassing less than 1\% of the reference map area. This means that the field is not too complex to predict and the method allows a significantly reduced sampling of the area. Moreover, we do not need to rely on a simulator to generate reference maps which requires time and expertise to evaluate. Rather than relying on simplistic signal propagation assumptions, GLIP adeptly estimates the intricate propagation dynamics in an urban environment, duly considering building characteristics. Future enhancements aim to refine the model architecture through the integration of more appropriate loss functions, adding residual networks, and in frequency and temporal dimensions thus augmenting the electromagnetic field reconstruction quality.
\section*{Acknowledgment}
The work has been funded by the Métropole Européenne de Lille (MEL). The French government’s Beyond5G initiative, which was sponsored as a component of the country’s future investment program and strategy for economic recovery, provided also partial funding for this project. Discussions in the COST action CA20120 INTERACT are also invaluable source of inspiration. 
\bibliographystyle{ieeetr}

\bibliography{ref.bib}

\begin{thebibliography}{10}

\bibitem{7338410}
A.~M. Niknejad, S.~Thyagarajan, E.~Alon, Y.~Wang, and C.~Hull, ``A circuit designer's guide to {5G} mm-{W}ave,'' in {\em 2015 IEEE Custom Integrated Circuits Conference (CICC)}, pp.~1--8, 2015.

\bibitem{RePEc}
P.~Ahokangas, M.~Matinmikko-Blue, S.~Yrjölä, and H.~Hämmäinen, ``Platform configurations for local and private {5G} networks in complex industrial multi-stakeholder ecosystems,'' {\em Telecommunications Policy}, vol.~45, no.~5, p.~S0308596121000331, 2021.

\bibitem{gajvsek2015electromagnetic}
P.~Gaj{\v{s}}ek, P.~Ravazzani, J.~Wiart, J.~Grellier, T.~Samaras, and G.~Thur{\'o}czy, ``Electromagnetic field exposure assessment in europe radiofrequency fields (10 {MHz}--6 {GHz}),'' {\em Journal of exposure science \& environmental epidemiology}, vol.~25, no.~1, pp.~37--44, 2015.

\bibitem{international2020guidelines}
{International Commission on Non-Ionizing Radiation Protection (ICNIRP)}, ``Guidelines for limiting exposure to electromagnetic fields (100 {KHz} to 300 {GHz}),'' {\em Health physics}, vol.~118, no.~5, pp.~483--524, 2020.

\bibitem{bailey2019synopsis}
W.~H. Bailey, R.~Bodemann, J.~Bushberg, C.-K. Chou, R.~Cleveland, A.~Faraone, K.~R. Foster, K.~E. Gettman, K.~Graf, T.~Harrington, {\em et~al.}, ``Synopsis of {IEEE} std c95. 1™-2019 “{IEEE} standard for safety levels with respect to human exposure to electric, magnetic, and electromagnetic fields, 0 {Hz} to 300 {GHz}”,'' {\em IEEE Access}, vol.~7, pp.~171346--171356, 2019.

\bibitem{matheron1963principles}
G.~Matheron, ``Principles of geostatistics,'' {\em Economic geology}, vol.~58, no.~8, pp.~1246--1266, 1963.

\bibitem{teganya2020data}
Y.~Teganya and D.~Romero, ``Data-driven spectrum cartography via deep completion autoencoders,'' in {\em ICC 2020-2020 IEEE International Conference on Communications (ICC)}, pp.~1--7, IEEE, 2020.

\bibitem{wang2020sensor}
S.~Wang and J.~Wiart, ``Sensor-aided {EMF} exposure assessments in an urban environment using artificial neural networks,'' {\em International Journal of Environmental Research and Public Health}, vol.~17, no.~9, p.~3052, 2020.

\bibitem{wang2022prediction}
S.~Wang, T.~Mazloum, and J.~Wiart, ``Prediction of {RF-EMF} exposure by outdoor drive test measurements,'' in {\em Telecom}, vol.~3, pp.~396--406, MDPI, 2022.

\bibitem{s20010311}
X.~Han, L.~Xue, F.~Shao, and Y.~Xu, ``A power spectrum maps estimation algorithm based on generative adversarial networks for underlay cognitive radio networks,'' {\em Sensors}, vol.~20, no.~1, 2020.

\bibitem{goodfellow2014generative}
I.~Goodfellow, J.~Pouget-Abadie, M.~Mirza, B.~Xu, D.~Warde-Farley, S.~Ozair, A.~Courville, and Y.~Bengio, ``Generative adversarial nets,'' {\em Advances in neural information processing systems}, vol.~27, 2014.

\bibitem{8794603}
Z.~Li, J.~Cao, H.~Wang, and M.~Zhao, ``Sparsely self-supervised generative adversarial nets for radio frequency estimation,'' {\em IEEE Journal on Selected Areas in Communications}, vol.~37, no.~11, pp.~2428--2442, 2019.

\bibitem{mallik2022towards}
M.~Mallik, A.~A. Tesfay, B.~Allaert, R.~Kassi, E.~Egea-Lopez, J.-M. Molina-Garcia-Pardo, J.~Wiart, D.~P. Gaillot, and L.~Clavier, ``Towards outdoor electromagnetic field exposure mapping generation using conditional {GAN}s,'' {\em Sensors}, vol.~22, no.~24, p.~9643, 2022.

\bibitem{mallik2022eme}
M.~Mallik, S.~Kharbech, T.~Mazloum, S.~Wang, J.~Wiart, D.~P. Gaillot, and L.~Clavier, ``{EME-Net}: A {U}-net-based indoor emf exposure map reconstruction method,'' in {\em 2022 16th European Conference on Antennas and Propagation (EuCAP)}, pp.~1--5, IEEE, 2022.

\bibitem{mallik2023eme}
M.~Mallik, B.~Allaert, A.~Tesfay, D.~P. Gaillot, J.~Wiart, and L.~Clavier, ``{EME-GAN}: A conditional generative adversarial network based indoor {EMF} exposure map reconstruction,'' in {\em 29{\^A} Colloque sur le traitement du signal et des image}, vol.~23, pp.~745--748, 2023.

\bibitem{ronneberger2021u}
O.~Ronneberger, P.~Fischer, and T.~Brox, ``{U-Net}: convolutional networks for biomedical image segmentation. arxiv150504597 cs. published online may 18, 2015,'' 2021.

\bibitem{mirza2014conditional}
M.~Mirza and S.~Osindero, ``Conditional generative adversarial nets,'' {\em arXiv preprint arXiv:1411.1784}, 2014.

\bibitem{egea2019vehicular}
E.~Egea-Lopez, F.~Losilla, J.~Pascual-Garcia, and J.~M. Molina-Garcia-Pardo, ``Vehicular networks simulation with realistic physics,'' {\em IEEE Access}, vol.~7, pp.~44021--44036, 2019.

\bibitem{amiot2013pylayers}
N.~Amiot, M.~Laaraiedh, and B.~Uguen, ``{Pylayers}: An open source dynamic simulator for indoor propagation and localization,'' in {\em 2013 IEEE International Conference on Communications Workshops (ICC)}, pp.~84--88, IEEE, 2013.

\bibitem{shi2014light}
L.~Shi, H.~Hassanieh, A.~Davis, D.~Katabi, and F.~Durand, ``Light field reconstruction using sparsity in the continuous fourier domain,'' {\em ACM Transactions on Graphics (TOG)}, vol.~34, no.~1, pp.~1--13, 2014.

\bibitem{daribo2010depth}
I.~Daribo and B.~Pesquet-Popescu, ``Depth-aided image inpainting for novel view synthesis,'' in {\em 2010 IEEE International workshop on multimedia signal processing}, pp.~167--170, IEEE, 2010.

\bibitem{long2019low}
Z.~Long, Y.~Liu, L.~Chen, and C.~Zhu, ``Low rank tensor completion for multiway visual data,'' {\em Signal processing}, vol.~155, pp.~301--316, 2019.

\bibitem{szczypkowski20243d}
P.~Szczypkowski, M.~Pawlowska, and R.~Lapkiewicz, ``{3D} super-resolution optical fluctuation imaging with temporal focusing two-photon excitation,'' 2024.

\bibitem{layer2015pet}
T.~Layer, M.~Blaickner, B.~Kn{\"a}usl, D.~Georg, J.~Neuwirth, R.~P. Baum, C.~Schuchardt, S.~Wiessalla, and G.~Matz, ``Pet image segmentation using a gaussian mixture model and markov random fields,'' {\em EJNMMI physics}, vol.~2, pp.~1--15, 2015.

\bibitem{buades2005review}
A.~Buades, B.~Coll, and J.-M. Morel, ``A review of image denoising algorithms, with a new one,'' {\em Multiscale modeling \& simulation}, vol.~4, no.~2, pp.~490--530, 2005.

\end{thebibliography}

\end{document}